\documentclass[lettersize,journal]{IEEEtran}

\usepackage[utf8]{inputenc} 
\usepackage[T1]{fontenc}    
\usepackage{hyperref}       
\usepackage{url}            
\usepackage{booktabs}       
\usepackage{amsfonts}       
\usepackage{nicefrac}       
\usepackage{microtype}      
\usepackage{xcolor}         
\usepackage{graphicx}
\usepackage{amsmath}
\usepackage{amsthm}
\usepackage[ruled,linesnumbered]{algorithm2e}
\usepackage{textcomp}
\usepackage{boldline}

\begin{document}

\title{Novel clustered federated learning
based on local loss}

\author{\normalsize
Endong Gu\thanks{Endong Gu is with the LMIB, School of Mathematical Sciences, Beihang University, Beijing 100191, P.R. China, e-mail: {\tt \color{blue}endonggu@buaa.edu.cn}},\
Yongxin Chen\thanks{Yongxin Chen is with the LMIB, School of Mathematical Sciences, Beihang University, Beijing 100191, P.R. China, e-mail: {\tt \color{blue}chenyongxin@buaa.edu.cn}},\
Hao Wen\thanks{Hao Wen is is with the College of Science, China Agricultural University, Beijing 100083, P.R. China, e-mail: {\tt \color{blue}wenh06@cau.edu.cn}},\
Xingju Cai\thanks{Xingju Cai is with the School of Mathematical Sciences,
Nanjing Normal University,
Nanjing 210023, P.R. China, e-mail: {\tt \color{blue}caixingju@njnu.edu.cn}}, \and
Deren Han\thanks{Deren Han is the corresponding author and is with the School of Mathematical Sciences, Beihang University, LMIB and NSLSCS, Beijing 100191, China, e-mail: {\tt \color{blue}handr@buaa.edu.cn}}

}


\newcommand{\urllib}{\url{https://github.com/wenh06/fl-sim}}
\newcommand{\urllcfl}{\url{https://github.com/wenh06/LCFL}}

\newtheorem{assumption}{Assumption}
\newtheorem{definition}{Definition}
\newtheorem{theorem}{Theorem}

\maketitle

\begin{abstract}
This paper proposes {\tt LCFL}, a novel clustering metric for evaluating clients' data distributions in federated learning. {\tt LCFL} aligns with federated learning requirements, accurately assessing client-to-client variations in data distribution. It offers advantages over existing clustered federated learning methods, addressing privacy concerns, improving applicability to non-convex models, and providing more accurate classification results. {\tt LCFL} does not require prior knowledge of clients' data distributions. We provide a rigorous mathematical analysis, demonstrating the correctness and feasibility of our framework. Numerical experiments with neural network instances highlight the superior performance of {\tt LCFL} over baselines on several clustered federated learning benchmarks.
\end{abstract}


\section{Introduction}
\label{sec:introduction}

The rapid development of the Internet and the mass popularization of private devices have generated vast amounts of personal data. Benefiting from the fast growth of storage and computational capacities, companies can fully utilize this data to predict market demand through machine learning methods and make profits. As a result, machine learning research has gradually shifted from model-driven to data-driven. Although the full mining of user data has brought more personalized services, people's concerns about their data being freely sold and privacy leakage have also followed. Various nations have progressively passed relevant laws and regulations on privacy protection \cite{Albrecht_2016}. In this context, a new machine learning paradigm known as "federated learning" has been proposed in the field of machine learning \cite{pmlr-v54,konevcny2016federated}. It investigates how to use data from all parties (which we refer to as clients) to complete machine learning while upholding user privacy.

\begin{figure}[htbp] 
    \centering
    \includegraphics[width=\linewidth]{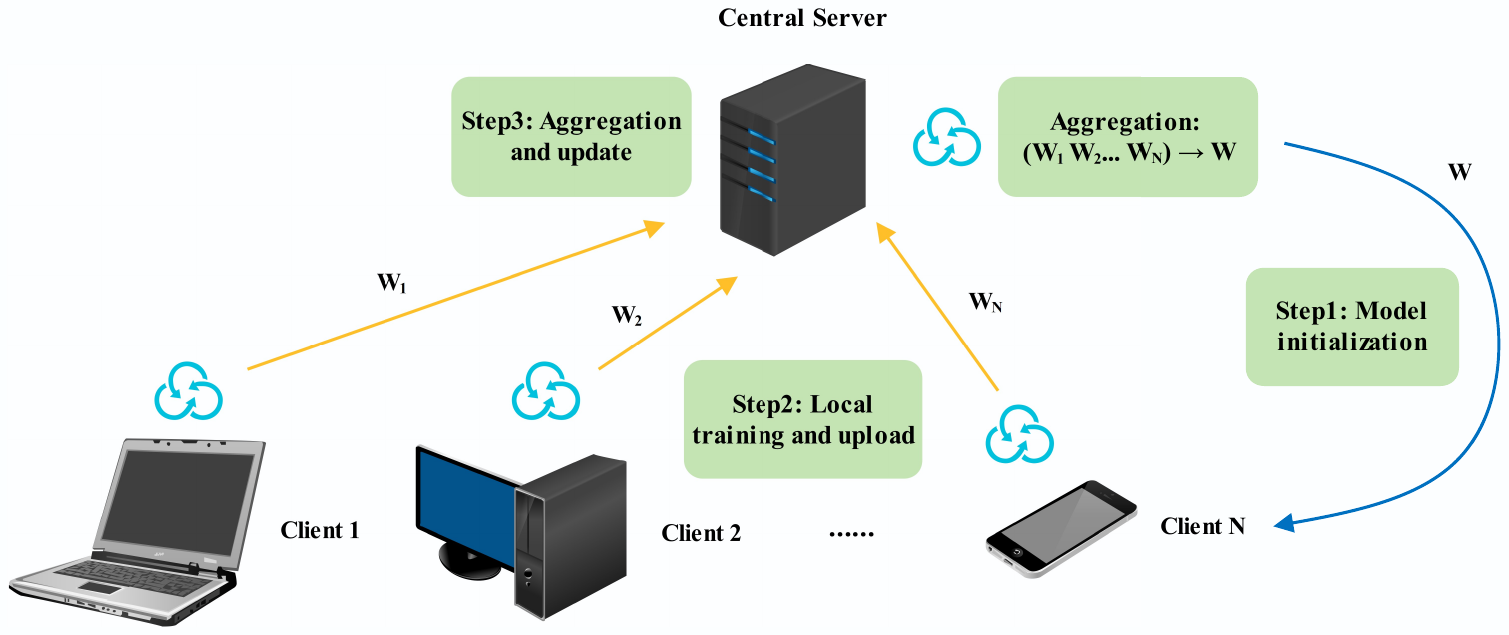}
    \caption{{\tt FL} three-step protocol illustration}
    \label{fig:fl three step protocol}
\end{figure}

The term "federated learning", abbreviated as {\tt FL}, was introduced in 2016 by Google's academic team \cite{pmlr-v54} which is a distributed machine learning model training strategy that focuses more on privacy protection and communication efficiency. A standard {\tt FL} progress includes an iterative three-step protocol \cite{lim2020federated} (illustrated in Figure \ref{fig:fl three step protocol}). In each communication round, denoted as $t$, the clients first download the latest global model, denoted as $w_t$, from the central server. This ensures that all clients start with the same initial model. Then each client independently improves the downloaded model by using their respective local data, denoted as $D_i$, to train the model on their devices. The local training process typically involves performing multiple iterations of an optimization algorithm, such as stochastic gradient descent, on mini-batches of data to prevent overfitting and enhance generalization. This process allows the model to learn from the unique characteristics of each client's data and valuable work focusing on this process has been done \cite{li2021ditto}. After local model training, all clients upload their updated model parameters to the central server. Finally, the server aggregates the received model updates. The aggregation step can involve simple averaging or more sophisticated techniques, such as secure aggregation or weighted averaging considering the importance of each client's update. This step ensures that the aggregated model reflects a collective representation of all clients' contributions while accounting for variations in the size and quality of local datasets.

By iteratively executing these steps across multiple rounds, {\tt FL} enables the collaborative training of a shared model while preserving data privacy. The protocol ensures that clients can benefit from the aggregated knowledge of the entire network without directly sharing their raw data with the server or other clients.

Two key challenges distinguish {\tt FL} from traditional distributed optimization: high degrees of systems and statistical heterogeneity \cite{MAL,CMFD,CA}. System heterogeneity in {\tt FL} encompasses significant variations in system characteristics among devices involved in the learning process, including disparities in hardware specifications (such as CPU and memory), network connectivity options (such as 3G, 4G, 5G, and Wi-Fi), and power availability (e.g., battery level). These diverse system-level attributes introduce additional challenges, particularly in the areas of straggler mitigation and fault tolerance within the federated network \cite{FedProx}. On the other hand, statistical heterogeneity in {\tt FL} refers to the existence of non-identically distributed data across the network, where each device or client possesses a distinct local dataset influenced by factors such as user preferences, geographical locations, or device-specific characteristics. This statistical heterogeneity gives rise to challenges concerning varying model performance and potential biases and fairness issues.



Federated Averaging ({\tt FedAvg}) is a fundamental algorithm in {\tt FL} that enables the training of machine learning models on decentralized data while preserving privacy. It involves iterative rounds of training on local devices, followed by aggregation of model updates through a central server using weighted averaging. However, {\tt FedAvg} has drawbacks, including high communication and bandwidth requirements, challenges with straggler devices that can slow down the training, and difficulties in handling non-IID data distributions. Ongoing research aims to address these limitations to enhance the efficiency and effectiveness of {\tt FL} \cite{li2021ditto,FedProx,FedSplit,FedPD,FedDR,hanzely2020federated}.

These works focus on improving efficiency and effectiveness with non-IID data \cite{pmlr-v54}. Federated Proximal ({\tt FedProx}) is an algorithm for federated learning that tackles this non-IID data challenge \cite{FedProx}. It extended {\tt FedAvg} by introducing a proximal term to the optimization objective to encourage consistent model updates across devices. It improved global model convergence and generalization performance. However, {\tt FedProx} has drawbacks including hyperparameter sensitivity, communication overhead, and lack of explicit handling of straggler devices. Splitting Scheme for Solving Federated Problems ({\tt FedSplit}) is an algorithmic framework introduced in \cite{FedSplit} for efficient distributed convex minimization in a hub-and-spoke model inspired by {\tt FL}. \cite{FedSplit} examined previous procedures, {\tt FedAvg} and {\tt FedProx}, revealing that their fixed points do not necessarily correspond to stationary points of the original optimization problem, even in simple convex scenarios with deterministic updates. To address this issue, {\tt FedSplit} applied the Peaceman-Rachford splitting technique \cite{peaceman1955numerical} into {\tt FL}, ensuring that fixed points align with the optima of the original problem. However, it should be noted that {\tt FedSplit} does not explicitly account for system heterogeneity, which limits its practicality in {\tt FL} settings. \cite{FedPD} also noticed the problem that {\tt FedAvg} and {\tt FedProx} cannot converge to the global stationary solution and proposed a novel algorithmic framework based on primal-dual optimization called Federated Primal-Dual Algorithm ({\tt FedPD}) to address this problem. Nevertheless, {\tt FedPD}'s requirement for all clients to participate in each communication round makes it less practical and applicable in {\tt FL} settings as well.
To overcome this limitation, \cite{FedDR} combined a nonconvex Douglas-Rachford splitting method \cite{douglas1956numerical}, randomized block-coordinate strategies, and asynchronous implementation and proposed Federated Douglas-Rachford ({\tt FedDR}). Unlike previous methods such as {\tt FedSplit} and {\tt FedPD}, {\tt FedDR} updates only a subset of users at each communication round, potentially in an asynchronous manner, making it more practical for real-world implementations. It also achieves the best-known $O(\epsilon^{-2})$ communication complexity for finding a stationary point under standard assumptions, where $\epsilon$ is a given accuracy.

Moreover, many researchers pay more attention to designing robust models \cite{pmlr-v108,pmlr-v97}, reducing expensive communication \cite{Barnes} and preserving the privacy of user data \cite{Patel,augenstein2019generative}. It is necessary to not only consider the similarity between data and integrate similar data to improve model performance but also identify the data with differences and develop model personalization ability. Personalized {\tt FL} that meets the two requirements at the same time has become one of the hot topics in {\tt FL} research \cite{MAL,CMFD}. To address statistical heterogeneity, clustered {\tt FL} is also a hot topic that is studied in this paper. For more details, the readers are referred to the survey papers \cite{MAL,CMFD,CA,TKDE}.

Before introducing clustered {\tt FL}, it is essential to provide a concise overview of clustering methods, a classical unsupervised machine learning technique. Unsupervised machine learning aims to explore and uncover the inherent structure and patterns within unlabeled training data, without relying on class label information. Clustering, as a fundamental task in unsupervised learning, partitions a dataset into disjoint subsets, called clusters, to identify potential categories. Given the vast array of clustering algorithms available, we will highlight some classical algorithms categorized into four main groups: partitioning-based, density-based, hierarchical-based, and model-based \cite{zhou2021machine}.

Partitioning-based algorithms rapidly determine clusters by initially assigning samples and iteratively reallocating them to appropriate groupings. One well-known algorithm in this category is $k$-means, which has been reinvented many times in history by scholars in various fields such as Steinhauar (1956), Lloyd (1957), and McQueen (1967) \cite{jain1988algorithms,jain2010data}. Variants such as $k$-medoids \cite{kaufmann1987clustering} enforce cluster centers to be actual training samples, while the number of clusters, $k$, is typically predetermined, although heuristics exist for automatic determination \cite{pelleg2000x,tibshirani2001estimating}. Density-based clustering algorithms characterize the sample distribution density to discover clusters of arbitrary shapes. Noteworthy methods in this category include Density-Based Spatial Clustering of Applications with Noise ({\tt DBSCAN}) \cite{ester1996density}, Ordering Points To Identify the Clustering Structure ({\tt OPTICS}) \cite{ankerst1999optics} and Density-Based Clustering of Applications with Noise using Local Updates ({\tt DENCLUE}) \cite{hinneburg98efficient}. Hierarchical-based algorithms organize data hierarchically based on proximity. Agglomerative Nesting ({\tt AGNES}) \cite{kaufman2009finding} adopts a bottom-up strategy, while Divisive Analysis ({\tt DIANA}) \cite{kaufman2009finding} follows a top-down approach. Hierarchical clustering algorithms like Balanced Iterative Reducing and Clustering using Hierarchies ({\tt BIRCH}) \cite{zhang1996birch} and Robust Clustering using Links ({\tt ROCK}) \cite{guha2000rock} improve upon {\tt AGNES} and {\tt DIANA} by allowing backtracking adjustments on merged or split clusters. Model-based methods optimize the fit between the given data and a predefined mathematical model, assuming the data is generated by a mixture of underlying probability distributions. Model-based Clustering ({\tt MCLUST}) \cite{fraley2002model} is a well-known algorithm in this category. For a more comprehensive understanding of clustering methods, we recommend referring to the papers \cite{jain1988algorithms,xu2005survey,gan2020data,fahad2014survey}.

Given that clustering can uncover potential patterns in data, it is a natural approach to cluster users with similar data distributions into cohesive groups and train a {\tt FL} model for each cluster \cite{Sattler,Ghosh2}. However, the challenge of clustering in {\tt FL} arises from the fact that the classification goal is based on the distribution of user data, which cannot be shared in real-world scenarios. Consequently, it becomes necessary to employ alternative information, rather than the distribution itself, to discern the differences among users.

In early research on {\tt FL}, the prevailing assumption was that all users shared a single global model. However, the heterogeneity of data distribution often led to sub-optimal performance of the global model on individual user data \cite{Sattler}. To address this statistical heterogeneity, personalized {\tt FL} approaches have emerged, including the notable clustered {\tt FL}. Clustering users based on data distribution has gained significant attention in various applications such as recommendation systems and precision medicines.

To estimate the clustering structure of data distribution, \cite{Sattler} employed the gradient cosine similarity $\alpha_{i,j}$ as a metric for user classification using the bisection method and hierarchical structures to select the optimal cluster.
\begin{equation*}
    \alpha_{i,j} = \frac {\langle\nabla L_i(w^\ast),\nabla L_j(w^\ast) \rangle}{\|\nabla L_i(w^\ast)\|\|\nabla L_j(w^\ast)\|}
\end{equation*}
Subsequently, a customized {\tt FL} model is provided for each cluster. However, it is important to note that \cite{Sattler} relies on gradient information, which poses inherent privacy risks \cite{Zhu}. Moreover, their clustering procedure is centralized, resulting in high computational costs, especially when dealing with a large number of devices.

To overcome these limitations, \cite{Ghosh1} and \cite{FPFC} utilize local model parameters trained by users instead of relying on gradient information as metrics \eqref{metric:model parameters}.
\begin{equation}\label{metric:model parameters}
    d_{i,j} = \|w_i^\ast - w_j^\ast\|
\end{equation}
\cite{FPFC} introduced non-convex paired penalty functions and proposed the Fusion Penalized Federated Clustering({\tt FPFC}) algorithm, which performs the clustering process and model updates simultaneously, eliminating the need for central clustering. However, we observe that employing model parameters as metrics may lead to incorrect classification and deviate from the original intent of clustering in federated learning.

Drawing inspiration from the classical $k$-means algorithm, Mansor et al. and Ghosh et al. proposed similar algorithms, namely HYPCLUSTER \cite{Mansor} and the Iterative Federated Clustering Algorithm ({\tt IFCA}) \cite{Ghosh2}, respectively. These algorithms alternate between the clustering and model update processes. The cluster assignment for each user is estimated by minimizing the loss function. Specifically, the cluster identity $\hat{j}$ of client $i$ is estimated by
\begin{equation*}
    \hat{j} = \arg\min_{j\in[k]}L_i(w_j).
\end{equation*}
\cite{Ghosh2} established the convergence rate of the population loss function under favorable initialization, ensuring both convergence of the training loss and generalization to test data. \cite{Mansor} provided guarantees only for generalization. While these algorithms eliminate the need for centralized clustering, they still require the specification of the number of clusters $k$, and the clustering results are typically sensitive to the initial values.

In our proposed clustering framework, we adopt a loss function-based metric \eqref{metric}, which avoids the privacy leakage caused by using gradient information.
\begin{equation}\label{metric}
    d(i,j) = |L_i(w_i) - L_i(w_j)| + |L_j(w_i) - L_j(w_j)|.
\end{equation}
Unlike \cite{Ghosh2} and \cite{Mansor}, our framework does not assume the cluster number $k$ in advance. Furthermore, our framework is flexible and compatible with various clustering methods, such as binary hierarchical clustering or density-based clustering algorithms like {\tt DBSCAN}. If the value of $k$ is known, the $k$-medoids clustering algorithm can also be utilized.

\paragraph{Contributions} In this paper, we present a novel clustering framework based on loss function metrics, called "Loss based Clustered Federated Learning ({\tt LCFL})". The primary objective of {\tt LCFL} is to enable the use of clustering methods in {\tt FL} while ensuring compliance with data protection regulations. Our main contributions can be summarized as follows:

\begin{enumerate}
    \item Flexibility in clustering methods: {\tt LCFL} does not impose any specific clustering algorithm, but instead supports partitioning-based clustering, density-based clustering, and hierarchical-based clustering methods. This flexibility allows practitioners to choose the most suitable clustering method based on their practical experience and specific requirements.

    \item Efficient and privacy-preserving solution: By leveraging loss function metrics, {\tt LCFL} provides an efficient and privacy-preserving solution for clustering in {\tt FL}. Instead of sharing raw data, our framework makes use of the loss function metrics, addressing the challenges of data privacy and security, allowing multiple clients to collaborate and contribute their local data without compromising individual data privacy.

    \item Valuable contribution to {\tt FL}: The {\tt LCFL} framework offers a valuable contribution to the field of {\tt FL} by facilitating the application of clustering algorithms in a privacy-preserving manner. Its compatibility with different clustering methods and {\tt FL} methods enhances its practicality and usefulness in various {\tt FL} scenarios.
\end{enumerate}
By combining these contributions, {\tt LCFL} empowers practitioners to utilize clustering methods effectively in {\tt FL} while upholding data protection norms and ensuring privacy preservation.

\paragraph{Notation} We use [M] to denote the set of integers $\{1, 2, \ldots, M\}.$ We denote by $X$ the set of all possible examples or instances which is also referred to as the input space. The set of all possible labels or target values is denoted by $Y.$ The data stored on the client $i$ is denoted by $X_i,$ the number of which is denoted by $m_i.$ We use $x_i$ to refer to a sample in the dataset $X_i.$ We assume that any sample $x_i \in X_i$ is independently and identically distributed according to a fixed but unknown distribution $D_i$. Let $f(x;w):X \rightarrow Y$ be the machine learning model associated with $w$, where $w\in\mathcal{W}$ is the model parameter to be learned from data. The loss function of a model $f(\cdot;w)$ on some example $x\in X$ also referred to as the risk or true error of $f$ is denoted by $l(f(x;w),y)$ (sometimes dismissing $w$ for simplicity). We use $L_i(w)$ to denote the empirical loss of model $f(\cdot;w)$ on data $X_i,$ that is $L_i(w) = \frac{1}{m_i}\sum_{k=1}^{m_i}l(f(x_{i,k};w),y_{i,k}).$ And the optimal value of $L_i(w)$ is denoted by $L_i^\ast,$ that is $L_i^\ast = \min_{w\in\mathcal{W}}L_i(w).$ If $(X,y)\sim D,\ \mu_X, \mu_y$ denote the expectation of $X,\ y$ separately, $\Sigma_{XX}$ denotes the variance of $X,\ \Sigma_{Xy}$ denotes the covariance of $X$ and $y,$ and so on.

The subsequent sections of this paper are organized as follows. Section \ref{sec:method} presents a description of our clustered FL framework, namely {\tt LCFL}, accompanied by the introduction of our main theorem. Additionally, implementation considerations regarding our framework are discussed. Supplementary explanations of our theorem are provided in Section \ref{sec:Examples_and_mathematical_analysis}. In Section \ref{section:drawbacks}, we outline the limitations associated with employing model parameters. Section \ref{sec:numerical_experiments} presents the experimental results we obtained, followed by the conclusion of our study in Section \ref{sec:conclusion}.

\section{Clustered federated learning based on local loss}
\label{sec:method}

Due to privacy concerns, the propagation of the user's data distribution is not feasible in clustered {\tt FL}. Therefore, an alternative approach is required to substitute for the data distribution information. Zhu et al. \cite{Zhu} have highlighted the privacy risks associated with using model gradient information, making gradient cosine similarity an unsuitable clustering metric. In Section \ref{section:drawbacks}, we elaborate on the limitations of using model parameters as a metric, particularly when the best model parameter is not unique, leading to incorrect classification results. While \cite{Ghosh2} demonstrated that it is possible to achieve satisfactory clustering performance without a metric function, it is necessary to have a good initial setup. Consequently, a new metric function is needed to classify clients and address the aforementioned scenario effectively.

\subsection{Algorithm and Main Results}
\label{subsec Algorithm and Main Results}

In this subsection, we propose our algorithm framework and some implementation considerations about our framework. The motivation of clustered {\tt FL} is to reduce the impact of non-IID data by clustering users with similar data distribution into one cluster. For users with very different local data distributions, model parameters trained using only local data will not work on other user data \cite{Mansor}. When $w_i^\ast\in \arg\min_{w\in\mathcal{W}}L_i(w),\ w_j^\ast\in \arg\min_{w\in\mathcal{W}}L_j(w),$ according to optimality, we have
\begin{equation*}
    \forall w\in\mathcal{W},\ L_i^\ast = L_i(w_i^\ast)\leq L_i(w).
\end{equation*}
If two users' datasets $X_i, X_j$ come from two distributions $D_i, D_j$ which we cannot check whether they are different, there will be two cases. The first we expect is
\begin{equation}\label{want}
    L_i(w_j^\ast) > L_i(w_i^\ast).
\end{equation}
Otherwise the case $L_i(w_j^\ast) = L_i(w_i^\ast) = L_i^\ast$ indicates that
\begin{equation*}
    w_j^\ast\in \arg\min_{w\in\mathcal{W}}L_i(w)
\end{equation*}
which means datasets $X_i, X_j$ share the same best model parameter $w_j.$ Then, if the two users collaborate on training one model $L_{i+j}(w) = \frac{m_i}{m_i+m_j}L_i(w) + \frac{m_j}{m_i+m_j}L_j(w)$ it will be:
\begin{equation*}
    \begin{aligned}
        L_{i+j}(w_j^\ast) & = \frac{m_i}{m_i+m_j}L_i^\ast + \frac{m_j}{m_i+m_j}L_j^\ast \\
        & \leq \frac{m_i}{m_i+m_j}L_i(w) + \frac{m_j}{m_i+m_j}L_j(w),\ \forall w\in\mathcal{W}.
    \end{aligned}
\end{equation*}
which means that $w_j^\ast$ is still the best model parameter. Under this case, both users $i,j$ will not get a worse model after cooperation. Moreover, because more data are used in the training process, the generalization performance of the model $L_{i+j}(w)$ will be improved. This result shows assigning user $i,j$ in the same cluster is appropriate.

When considering the first case mentioned in equation \eqref{want}, we assert that the two distributions, $D_i$ and $D_j$, exhibit distinct characteristics. By "different", we mean that these distributions can be discerned by the model hypothesis set $\mathcal{W}$. In the subsequent context of this section, we will further demonstrate and provide evidence for this claim. Drawing inspiration from the aforementioned analyses and the inherent connection between the loss function and distribution, we propose utilizing equation \eqref{metric} as the clustering metric. In Section \ref{sec:Examples_and_mathematical_analysis}, we present an illustrative example to illustrate the correlation between the loss function and distribution.

We utilize equation \eqref{metric} as a measure of the "distance" between the data distributions of users. This metric function, by definition, is symmetric and satisfies the triangle inequality for any loss function $L$. While the gradient cosine similarity $\alpha_{i,j}$ exhibits symmetry, the model parameter metric $d_{i,j}$ possesses both symmetry and satisfies the triangle inequality. However, it is important to note that the expression \eqref{metric} may not always qualify as a true distance. This is because there may exist cases where $d(i,j) = 0$ for $D_i \neq D_j$, which violates the requirement of a distance metric. The same applies to the other two metrics; they are not actual distances either.

We provide a companion algorithmic framework for using this metric function in {\tt FL}, taking into account both communication and practical considerations. The main idea of the algorithm is to cluster the users before conducting {\tt FL} and apply the {\tt FL} algorithm to each cluster separately. We do not require each user to communicate after achieving the optimal solution locally. Instead, we use a parameter called the local iteration count $T$. The local training process ends when the local iteration count reaches the desired value. This preliminary step can be seen as a warm-up phase for {\tt FL}. The algorithm is formally presented in Algorithm \ref{algorithm}.

\begin{algorithm}
\caption{Loss based Clustered Federated Learning(LCFL)}\label{algorithm}
\KwIn{initialization $w_i^0,i\in[M],$ number of local iterations $T,$ local step size $\gamma,$ clustering method $C(\cdot)$, federated learning method FedOpt}
\For{Client $i\in[M]$ in parallel}{
\For{$t = 0$ \KwTo $T-1$}{
$w_i^{t+1} = w_i^t - \gamma \nabla L_i(w_i^t);$}
Sends $w_i^T$ to Server\;
\For{$j\in[M]/i$}{
Client $i$ receives $w_j^T$ from Server\;
$d(i,j)_i = \|L_i(w_j^T) - L_i(w_i^T)\|$\;
Sends $d(i,j)_i$ to Server;}}
\For{$i,j\in[M],i\neq j,$ Server}{
$d(i,j) = d(i,j)_i + d(j,i)_j$}
Server uses $C(d(i,j))$ to obtain clustering structure $C_1,\ldots,C_k$\;
\For {Client $k$ in clustering structure $C_k$ }{
$w_k^\ast = $FedOpt($C_k$).}
\KwOut{$C_1,\ldots,C_k,\ w_1^\ast,\ldots,w_k^\ast$}
\end{algorithm}

In the framework of the above clustering distance function $d(i,j)$, the $k$-means algorithm cannot be used, because in this framework, there are only distances between any two participants, and there is no way to choose the "cluster center", that is, the clustering method that requires the identity information of participants is not applicable in our clustering framework. However, a distance metric alone is sufficient for most algorithms, such as improvements to $k$-means. This $k$-medoids method selects the medoid based only on distance, and most hierarchical and density-based clustering methods.

Now we turn to prove that if $d(i,j) \geq R,$ distributions $D_i, D_j$ can be distinguished by the model hypothesis set $\mathcal{W}.$ We will prove this property starting with the measure of the distribution. The divergence between distributions is usually measured by Kullback-Leibler divergence (KL divergence) \cite{kullback1951information}.
\begin{definition}[KL divergence \cite{kullback1951information}]
    For discrete probability distributions $P$ and $Q$ defined on the same sample space $X$ the KL divergence from $Q$ to $P$ is defined as:
    \begin{equation*}
        D_{KL}(P||Q) = \sum_{x\in X} P(x)\log(\frac{P(x)}{Q(x)}).
    \end{equation*}
\end{definition}
However, the KL divergence is only related to the distributions and does not consider the machine learning task that this paper cares about most. To measure the difference in data distributions under a machine learning model, we use the concept of "label-discrepancy" between distributions in \cite{mohri2012new}:
\begin{definition}[Label-discrepancy \cite{mohri2012new}]
    Given a loss function $L$, the discrepancy $disc$ between two distributions $D_i$ and $D_j$ over $X\times Y$ is defined by:
    \begin{equation*}
        disc_\mathcal{W}(D_i,D_j) = \sup_{w\in\mathcal{W}}|L_{D_i}(w) - L_{D_j}(w)|,
    \end{equation*}
    where $D_i,D_j$ is the distribution of  the feature-label data, $\mathcal{W}$ is the hypothesis set for machine learning models, $L_{D_i}(w)$ denotes the expected loss function when the model is $w$ and the data distribution is $D_i,$ that is $L_{D_i}(w) = \mathbb{E}_{(x,y)\sim D_i}l(f(x;w),y).$
\end{definition}

Under this definition, two distributions $D_i$ and $D_j$ are distinguishable under $\mathcal{W}$ if and only if the label discrepancy between them is $0$. If $D_i = D_j$, it is evident that $disc_\mathcal{W}(D_i,D_j)=0$. Conversely, if $disc_\mathcal{W}(D_i,D_j)=0$, it implies that for every $w\in\mathcal{W}$, $L_{D_i}(w) = L_{D_j}(w)$. In this case, the two distributions are not distinguishable under $\mathcal{W}$. This means that the loss of all models in the hypothesis set is the same under both $D_i$ and $D_j$, indicating that models trained on $D_i$ generalize well on $D_j$ and vice versa. Moreover, it is advisable to assign users $i$ and $j$ to the same cluster based on this observation.

We have observed that the expected form of $d(i,j)$, denoted as $\hat{d}(i,j)$, is less than $2\cdot disc_\mathcal{W}(D_i,D_j)$:
\begin{equation*}
    \begin{aligned}
        \hat{d}(i,j) & = |L_{D_i}(w_i^\ast) - L_{D_i}(w_j^\ast)| + |L_{D_j}(w_i^\ast) - L_{D_j}(w_j^\ast)| \\
        & \leq 2\sup_{w\in\mathcal{W}}|L_{D_i}(w) - L_{D_j}(w)| \\
        & = 2\cdot disc_\mathcal{W}(D_i,D_j).
    \end{aligned}
\end{equation*}
If there exists $\epsilon > 0$ such that $\hat{d}(i,j) \geq \epsilon$, we can conclude
\begin{equation}\label{case:geq0}
    0 < \frac{\epsilon}{2} \leq \frac{1}{2}\hat{d}(i,j) \leq disc_\mathcal{W}(D_i,D_j).
\end{equation}

Based on the previous analysis, we can deduce that under \eqref{case:geq0}, models from $\mathcal{W}$ cannot distinguish between the two distributions $D_i$ and $D_j$. Furthermore, the two distributions should be grouped into the same cluster.

However, $d(i,j)$ and $\hat{d}(i,j)$ are not equal in general. To analyze the bound of the two terms, we first introduce the definitions of empirical and average Rademacher complexity \cite{mohri2018foundations}. The formal definitions of the empirical and average Rademacher complexity are stated as follows.

\begin{definition}[Empirical Rademacher complexity \cite{mohri2018foundations}]
    Let $\mathcal{G}$ be a family of functions mapping from $Z$ to $[a,b]$ and $S = (z_1,\ldots,z_m)$ be a fixed sample of size $m$ with elements in $Z$. Then, the empirical Rademacher complexity of $\mathcal{G}$ with respect to the sample $S$ is defined as:
    \begin{equation*}
        \hat{\mathcal{R}}_S(\mathcal{G}) = \mathop{\mathbb{E}}_\sigma \left[ \sup_{g\in\mathcal{G}}\frac{1}{m}\sum_{i=1}^m\sigma_ig(z_i) \right],
    \end{equation*}
    where $\sigma = (\sigma_1,\ldots,\sigma_m)^\top$, with $\sigma_i$ is independent uniform random variables taking values in $\{-1,\ +1\}$. The random variables $\sigma_i$ are called Rademacher variables.
\end{definition}

\begin{definition}[Rademacher complexity \cite{mohri2018foundations}]
    Let $D$ denote the distribution according to which samples are drawn. For any integer $m\geq1$, the Rademacher complexity of $\mathcal{G}$ is the expectation of the empirical Rademacher complexity over all samples of size $m$ drawn according to $D$:
    \begin{equation*}
        \mathcal{R}_m(\mathcal{G}) = \mathop{\mathbb{E}}_{S\sim D^m}\left[ \hat{\mathcal{R}}_S(\mathcal{G}) \right].
    \end{equation*}
\end{definition}

The Rademacher complexity measures the richness of a family of functions by quantifying the extent to which the hypothesis set can fit random noise. For further details on the Rademacher complexity, we refer readers to the book by Mohri et al. \cite{mohri2018foundations}.

These definitions are provided in the general case of a family of functions $\mathcal{G}$ mapping from an arbitrary input space $Z$ to $\mathbb{R}$. However, under Assumption \ref{ass:bounded}, we can treat an arbitrary loss function $l:Y\times Y \rightarrow \mathbb{R}$ as $\mathcal{G}$, where $Z = X\times Y$:
\begin{equation*}
    \mathcal{G} = \{ g:(x,y)\rightarrow l(f(x;w),y) | w\in\mathcal{W}\}.
\end{equation*}

\begin{assumption}\label{ass:bounded}
    The loss function $l$ is bounded by some $M > 0$, that is $l(y, y') \leq M$ for all $y, y'\in Y$ or, more strictly, $l(f(x;w), y) \leq M$ for all $w \in\mathcal{W}$ and $(x, y)\in X \times Y$, the problem is referred to as a bounded problem.
\end{assumption}

Additionally, we assume that the range of the loss function $l$ is $[0,1]$, so we can omit the explicit mention of $M$ in the following theorem for brevity.

Now, armed with the definitions of empirical and average Rademacher complexity, we can proceed with the analysis of the term to be bounded and give the following theorem.

\begin{theorem}\label{thm:bound}
    For any $\delta > 0,$ with probability at least $(1-\delta)^4$ over the draw of $i.i.d$ samples $X_i, X_j$ of sizes $m_i, m_j$ respectively, the following holds:
    \begin{equation}\label{desire}
        \begin{aligned}
            |d(i,j)-\hat{d}(i,j)| \leq & 2\sqrt{\frac{\log(2/\delta)}{2m_i}} +  2\sqrt{\frac{\log(2/\delta)}{2m_j}} \\
            & + \mathcal{R}_{m_i}(\mathcal{W}) + \mathcal{R}_{m_j}(\mathcal{W}).
        \end{aligned}
    \end{equation}
\end{theorem}

We denote
\begin{equation*}
    \begin{aligned}
        C_\delta(X_i,X_j;\mathcal{W}) = & 2\sqrt{\frac{\log(2/\delta)}{2m_i}} +  2\sqrt{\frac{\log(2/\delta)}{2m_j}} \\
        & + \mathcal{R}_{m_i}(\mathcal{W}) + \mathcal{R}_{m_j}(\mathcal{W}).
    \end{aligned}
\end{equation*}
In the context of the Theorem \ref{thm:bound}, if the empirical discrepancy $d(i,j)$ between users $i$ and $j$ is greater than the bound $C_\delta(X_i, X_j;\mathcal{W}),$
then with probability at least $(1-\delta)^4,$ we obtain:
\begin{equation*}
    \hat{d}(i,j) \geq d(i,j) - C_\delta(X_i,X_j;\mathcal{W}) > 0
\end{equation*}
which indicates that users $i$ and $j$ should be assigned to the same cluster.

In simpler terms, if the observed difference between two users exceeds a certain threshold, the theorem guarantees that the expected difference will also be sufficiently large. This ensures that the clustering algorithm will correctly identify these users as belonging to the same cluster with a high level of confidence.

This result provides a statistical assurance for the performance of the clustering algorithm and the accuracy of its clustering decisions based on pairwise user distances.

Calculating the exact value of the Rademacher complexity can pose significant challenges. In fact, for certain hypothesis sets, computing the empirical Rademacher complexity is known to be NP-hard \cite{mohri2018foundations}. As a viable alternative, the VC dimension (Vapnik-Chervonenkis dimension \cite{1971Uniform}) is frequently employed to provide an upper bound on the Rademacher complexity. While we won't delve into the specific definition of the VC dimension in this article, it is widely utilized in machine learning theory to estimate the complexity of hypothesis classes and analyze the behavior of learning algorithms.


\subsection{Implementation considerations}
\label{subsec Implementation considerations}

In this subsection, we delve into the practical implementation details of our method, highlighting its seamless integration with the existing communication protocol of {\tt FL}. Furthermore, we will emphasize that our method maintains the privacy of the participating clients throughout the process.

The choice of the metric function $d(i,j)$ in clustering for {\tt FL} also takes into account the communication cost involved. In this context, it is important to minimize the amount of communication required between the participating clients.

One possible metric function is given by \eqref{badmetric}:
\begin{equation}\label{badmetric}
    \tilde{d}(i,j) = |L_i(w_i) - L_j(w_i)| + |L_i(w_j) - L_j(w_j)|,
\end{equation}
where $\tilde{d}(i,j)$ is defined as the difference between the local loss functions of clients $i$ and $j$ under their respective local models. Since exchanging local data is generally prohibited in {\tt FL} due to privacy concerns, calculating $|L_i(w_i) - L_j(w_i)|$ would require an additional communication step to transmit $L_i(w_i)$.

To address this issue and reduce the communication cost, the metric function $d(i,j)$ (as defined in our algorithm) is preferable. It relies on exchanging model parameters rather than local data. By exchanging the model parameters $w_i$ and $w_j$ between clients, the computation of $d(i,j)$ can be performed locally at each client without the need for additional data exchange. This approach reduces the communication overhead while still allowing the calculation of the pairwise distances necessary for clustering. Considering the goal of minimizing communication cost in {\tt FL}, the original metric function $d(i,j)$ is a more suitable choice compared to $\tilde{d}(i,j)$, as the former avoids the need for additional data communication.

Our algorithm acts as a warm-up phase before starting the {\tt FL} process, seamlessly integrating with existing {\tt FL} frameworks and systems without requiring significant modifications. It follows the standard {\tt FL} communication protocol, ensuring compatibility and easy adoption in various {\tt FL} setups. During the warm-up phase, our algorithm establishes a clustering structure among the clients. The local model parameters used for clustering training can also serve as initialization parameters for subsequent {\tt FL}. This utilization of pre-trained parameters speeds up the {\tt FL} process, as it provides a well-suited starting point. Thus, any communication costs during the warm-up stage are offset by the subsequent acceleration achieved through the use of pre-trained parameters.

Moreover, privacy preservation holds paramount importance in distributed learning scenarios. With our approach, we place a strong emphasis on protecting the privacy of the participating clients and ensuring the security and confidentiality of their sensitive data. One key aspect of our method is that it avoids the use of gradient information, which can be prone to information leakage.

In our algorithm, we do not explicitly check whether the discrepancy $d(\cdot,\cdot)$ between users $i$ and $j$ exceeds the specified threshold $C_\delta(i,j;\mathcal{W})$ to avoid the calculation of Rademacher complexity. Instead, we utilize the discrepancy matrix obtained from the clustering algorithm to derive the clustering structure. By employing the discrepancy matrix, we leverage the pairwise discrepancies between users to determine their similarities in the clustering process. This approach provides flexibility in capturing the inherent structure and relationships among the users, allowing the clustering algorithm to autonomously identify and group users based on the discrepancy matrix.

\section{Examples and mathematical analysis}
\label{sec:Examples_and_mathematical_analysis}

In this section, we provide an example to showcase the effectiveness of our algorithm when dealing with user data that exhibits different data distributions. By presenting this example, we aim to illustrate how our algorithm can successfully handle scenarios where the data distribution among users varies. This capability is crucial in clustered {\tt FL}, as it allows us to leverage diverse and distributed data sources while maintaining robust and accurate model performance.

We analyze our proposed algorithm in a concrete linear model. As mentioned earlier, let us denote the input space as $X$, and the set of target values as $y$. We assume that the data on a certain client are generated from a distribution of $D.$ Furthermore, we use the expected squared loss function $L(f(X;w),y) = \mathbb{E}_{(X,y)\sim D} \left(w^\top X - y\right)^2$ instead of the empirical form for the sake of theoretical analysis. The optimizer of this linear model is
\begin{equation*}
    w^\ast =(\Sigma_{XX}+\mu_X\mu_X^\top)^{-1}(\Sigma_{Xy}+\mu_y\mu_X),
\end{equation*}
and thus
\begin{equation}\label{eq:gl}
    \begin{aligned}
        & L(f(X;\hat{w}),y) - L(f(X;w^\ast),y) \\
        = & (\hat{w}-w^\ast)^\top\left[(\Sigma_{XX}+\mu_X\mu_X^\top)(\hat{w}+w^\ast)-2(\Sigma_{Xy}+\mu_y\mu_X)\right] \\
        \geq & \frac{\lambda_{\min}}{(\hat{\lambda}_{\max}+\hat{\mu}_X^\top\hat{\mu}_X)^2}\|\hat{\Sigma}_{Xy}+\hat{\mu}_X\hat{\mu}_y\|^2 \\
        & + \frac{1}{\lambda_{\max}+\mu_X^\top\mu_X} \|\Sigma_{Xy}+\mu_X\mu_y\|^2 \\
        & - \frac{2}{\hat{\lambda}_{\max}+\hat{\mu}_X^\top\hat{\mu}_X}\|\Sigma_{Xy}+\mu_X\mu_y\|\|\hat{\Sigma}_{Xy}+\hat{\mu}_X\hat{\mu}_y\|,
    \end{aligned}
\end{equation}
where $\hat{w}$ is the optimizer of this linear model on another data distribution $(X,y)\sim\hat{D},$ $\lambda_{\min}$ denotes the minimum eigenvalue of $\Sigma_{XX},$ $\lambda_{\max}$ denotes the maximum eigenvalue of $\Sigma_{XX}$ and $\hat{\lambda}_{\max}$ denotes the maximum eigenvalue of $\hat{\Sigma}_{XX}.$

If the condition
\begin{equation*}
    \|\Sigma_{Xy}+\mu_X\mu_y\| \leq \frac{\|\hat{\Sigma}_{Xy}+\hat{\mu}_X\hat{\mu}_y\|}{2(\hat{\lambda}_{\max} + \hat{\mu}_X^\top\hat{\mu}_X)}
\end{equation*}
is satisfied, then
\begin{equation*}
    \begin{aligned}
        & \frac{\lambda_{\min}}{(\hat{\lambda}_{\max}+\hat{\mu}_X^\top\hat{\mu}_X)^2}\|\hat{\Sigma}_{Xy}+\hat{\mu}_X\hat{\mu}_y\|^2 \\
        & + \frac{1}{\lambda_{\max}+\mu_X^\top\mu_X} \|\Sigma_{Xy}+\mu_X\mu_y\|^2 \\
        > & \frac{2}{\hat{\lambda}_{\max}+\hat{\mu}_X^\top\hat{\mu}_X}\|\Sigma_{Xy}+\mu_X\mu_y\|\|\hat{\Sigma}_{Xy}+\hat{\mu}_X\hat{\mu}_y\|
    \end{aligned}
\end{equation*}
holds. It means that the most right-hand side of \eqref{eq:gl} is greater than zero, hence the gap between $L(f(X;\hat{w}),y)$ and $L(f(X;w^\ast),y)$ exits. Accordingly using our proposed algorithm, one can obtain a clear distance matrix that is easy for clustering.

Moreover, we assume that all clients' data are normalized, such that $\mu_X = \hat{\mu}_X = \boldsymbol{0}$, $\mu_y = \hat{\mu}_y = 0$, and $\Sigma_{XX} = \hat{\Sigma}_{XX}$. This implies that the difference between distributions $D$ and $\hat{D}$ only exists in $\Sigma_{Xy}$. From this, we can derive the following expression:
\begin{equation*}
    \begin{aligned}
        & L(f(X;\hat{w}),y) - L(f(X;w^\ast),y) \\
        = & (\hat{\Sigma}_{Xy} - \Sigma_{Xy})^\top\Sigma_{XX}^{-1}(\hat{\Sigma}_{Xy} - \Sigma_{Xy}).
    \end{aligned}
\end{equation*}
Consequently, we can bound this term as follows:
\begin{equation}\label{eq:distribution and loss func}
    \begin{aligned}
        \frac{1}{\lambda_{\max}}\|\hat{\Sigma}_{Xy} - \Sigma_{Xy}\|^2 & \leq L(f(X;\hat{w}),y) - L(f(X;w^\ast),y) \\
        & \leq \frac{1}{\lambda_{\min}}\|\hat{\Sigma}_{Xy} - \Sigma_{Xy}\|^2.
    \end{aligned}
\end{equation}
The above inequality \eqref{eq:distribution and loss func} demonstrates the close connection between the loss function and the data distribution. This motivates us to consider equation \eqref{metric} as a suitable metric.

\section{Drawbacks of the model parameter metric}\label{section:drawbacks}

In this section, we elaborate that using model parameters as a metric will lead to wrong classification while using model loss functions will not.

The literature \cite{Ghosh1} states that using the model parameter $w^\ast$ as the metric of clustering needs to satisfy the following assumption:
\begin{assumption}
    \label{ass:model parameter}
    Model parameters of different clusters $w_i^\ast,w_j^\ast$ satisfy:
    \begin{equation*}
        \|w_i^\ast - w_j^\ast\| \geq R.
    \end{equation*}
\end{assumption}
However, we illustrate that models trained with the same data distribution can also satisfy the above separation assumption, which will result in users with the same data distribution being assigned to different clusters, which goes against the original intention of integrating similar data to improve model performance.

Consider using the Softmax model in {\tt FL}. Changing the Softmax model parameter $w \in \mathbb{R}^K$ to $w - \varphi = (w_1 - \varphi,\ldots,w_K - \varphi)^\top$ does not affect the model, since:
\begin{equation}
    \label{eq:Softmax}
    \begin{aligned}
        p(y_i = j|x_i;w-\varphi) & = \frac{e^{(w_j-\varphi)^T x_i}}{\sum_{l=1}^K e^{(w_l-\varphi)^T x_i}} \\
        & = \frac{e^{w_j^T x_i}e^{-\varphi^T x_i}}{\sum_{l=1}^K e^{w_l^T x_i}e^{-\varphi^T x_i}} \\
        & = \frac{e^{w_j^T x_i}}{\sum_{l=1}^K e^{w_l^T x_i}} \\
        & = p(y_i = j|x_i;w).
    \end{aligned}
\end{equation}

\eqref{eq:Softmax} means the Softmax model will get the same output under a family of model parameters: $\{w - \varphi|\varphi\in\mathbb{R}\}$. In this situation, even though the model parameters $w^\ast$ and $\hat{w}^\ast = w^\ast + \varphi$ are trained with the same data, they still satisfy the Assumption \ref{ass:model parameter}, for any $\varphi$ that satisfies the condition \eqref{eq:model parameter ass}.
\begin{equation}\label{eq:model parameter ass}
    \|w^\ast - \hat{w}^\ast\| = \|(\varphi,\ldots,\varphi)^\top\| \geq R.
\end{equation}
This means the model parameter metric divides users belonging to one cluster into two different clusters, resulting in the loss of data volume under this cluster.

Moreover, considering the Softmax model with regularization term $\|\cdot\|_2$ which is more commonly used in the field of machine learning, if
\begin{equation*}
    \varphi = \frac{2\sum_{k=1}^{K} w_k^\ast}{K},
\end{equation*}
such that both $w^\ast$ and $\hat{w}^\ast = w^\ast + \varphi$ are optimal solutions to the regularized Softmax model. In this situation, to cluster users with similar data distribution, it should have at least:
\begin{equation*}
    R > \|\varphi\| = \|\frac{2\sum_{k=1}^{K} w_k^\ast}{K}\|.
\end{equation*}
This implies that the choice of the $R$ value significantly impacts the clustering performance. In a broader context, when dealing with models that do not have a unique optimal solution (as is often the case with deep neural network models due to their combinatorial symmetry), it is not suitable to rely on the norm of model parameter differences as the clustering metric.

However, using our proposed metric function, this situation does not occur. The loss function commonly used for training a softmax model is the categorical cross-entropy loss. Mathematically, if we denote the predicted probabilities as $p$ and the true label as $y$ (both vectors of length equal to the number of classes $K$), the categorical cross-entropy loss can be computed as follows:
\begin{equation}\label{eq:Softmax loss}
    L = - \sum_{i=1}^{K} \left( y_i\log(p_i) \right).
\end{equation}
When the same dataset is used, according to \eqref{eq:Softmax}, it is obvious that the value of \eqref{eq:Softmax loss} will be the same. It is precisely this property that motivates us to use \eqref{metric} as a metric function.

\section{Numerical experiments}
\label{sec:numerical_experiments}

In this section, we present the experimental results that validate our theoretical analysis. We demonstrate the performance of our proposed metric, as well as other clustering metrics such as the difference of model parameters and the cosine of gradients. Additionally, we compare the performance of our algorithm with {\tt IFCA} and several baseline methods.

(\textbf{Machine}) We carried out the simulation experiments on a commodity workstation with one Intel{\textsuperscript{\tiny\textregistered}} Xeon{\textsuperscript{\tiny\textregistered}} Gold 5218R CPU with 64 GB RAM and 2 NVidia{\textsuperscript{\tiny\textregistered}} 3090 GPUs.

(\textbf{Implementation and Hyperparameters}) In order to conduct fair comparisons, Stochastic Gradient Descent ({\tt SGD}) was adopted as the local problem solver for all {\tt FL} algorithms involved in this paper. For all numerical experiments, the learning rates on the clients decay every global iteration by a factor of $0.99.$ The rest of the primary hyperparameters are summarized in Table \ref{tab:hyperparameters}. In this table, "participation rate" refers to the proportion of clients that participate in the training process every global iteration, and "init. learning rate" means the initial learning rates of the clients which decay with the pattern stated previously.

\begin{table}[htbp]
\centering
\caption{Primary hyperparameters of the numerical experiments conducted in this paper}
\label{tab:hyperparameters}
\begin{tabular}{|c|c|c|c|}
\hlineB{3.5}
hyperparameter & FEMNIST & Rotated MNIST & Rotated CIFAR10 \\
\hline \hline
participation rate & $30\%$ & $100\%$ & $10\%$ \\
init. learning rate & $0.03$ & $0.02$ & $0.02$ \\
batch size & $20$ & $20$ & $20$ \\
local epochs & $5$ & $10$ & $5$ \\
global iterations & $100$ & $60$ & $200$ \\
\hlineB{3.5}
\end{tabular}
\end{table}

The code, including the re-implementations of the {\tt FedAvg} and {\tt IFCA} algorithm, was built on top of a simulation framework {\tt fl-sim} for {\tt FL} based on PyTorch \cite{pytorch}. More details can be found in \href{https://github.com/wenh06/fl-sim}{https://github.com/wenh06/fl-sim} and \href{https://github.com/wenh06/LCFL}{https://github.com/wenh06/LCFL}.

We use three federated datasets to verify the performance: Federated Extended MNIST(FEMNIST) \cite{caldas2018_leaf}, Rotated MNIST and Rotated CIFAR10 \cite{Ghosh2}.

In the case of the FEMNIST dataset, we utilized a subsampled and repartitioned version of the original full dataset to introduce additional statistical heterogeneity. This dataset has also been employed in previous studies such as \cite{FedProx, FedDR}. Specifically, it followed the procedure of selecting 10 lowercase characters ('a'-'j') from the EMNIST dataset \cite{cohen2017emnist} and distributing only 3 classes to each device. The dataset consists of 200 devices, and we used a convolutional neural network model to classify the inputs, which were gray-scale images with 784 pixels and flattened to 1-dimensional.

The Rotated MNIST and Rotated CIFAR10 datasets were created based on the MNIST \cite{Lecun_1998_mnist} and CIFAR10 datasets \cite{cifar} by \cite{Ghosh2}. These datasets were designed to simulate a {\tt FL} scenario where the data distribution varies among different clients. For Rotated MNIST, the data pictures were rotated by four degrees: 0, 90, 180, and 270, resulting in the creation of $k=4$ clusters with a relatively straightforward cluster structure. Considering that the original MNIST dataset contains 60,000 training images and 10,000 test images, we randomly partitioned the 60,000$k$ training images and 10,000$k$ test images among $m$ clients, ensuring each client had the same rotation. Rotated CIFAR10 was created similarly to Rotated MNIST, except that the rotation degrees were 0 and 180, resulting in $k=2$ clusters, and we set the number of clients to $m=200$, as in \cite{Ghosh2}.

\subsection{Performances of the metrics}

In this subsection, we compared the performance of three different metrics which use model loss functions, model parameters, and the cosine of gradients. We utilized our proposed framework, using the same federated optimizer {\tt FedAvg} for fairness. However, the metrics used in the clustering stage are different. We performed experiments on two datasets: FEMNIST and Rotated MNIST.

\begin{figure}[htbp] 
   \centering
   \includegraphics[width=\linewidth]{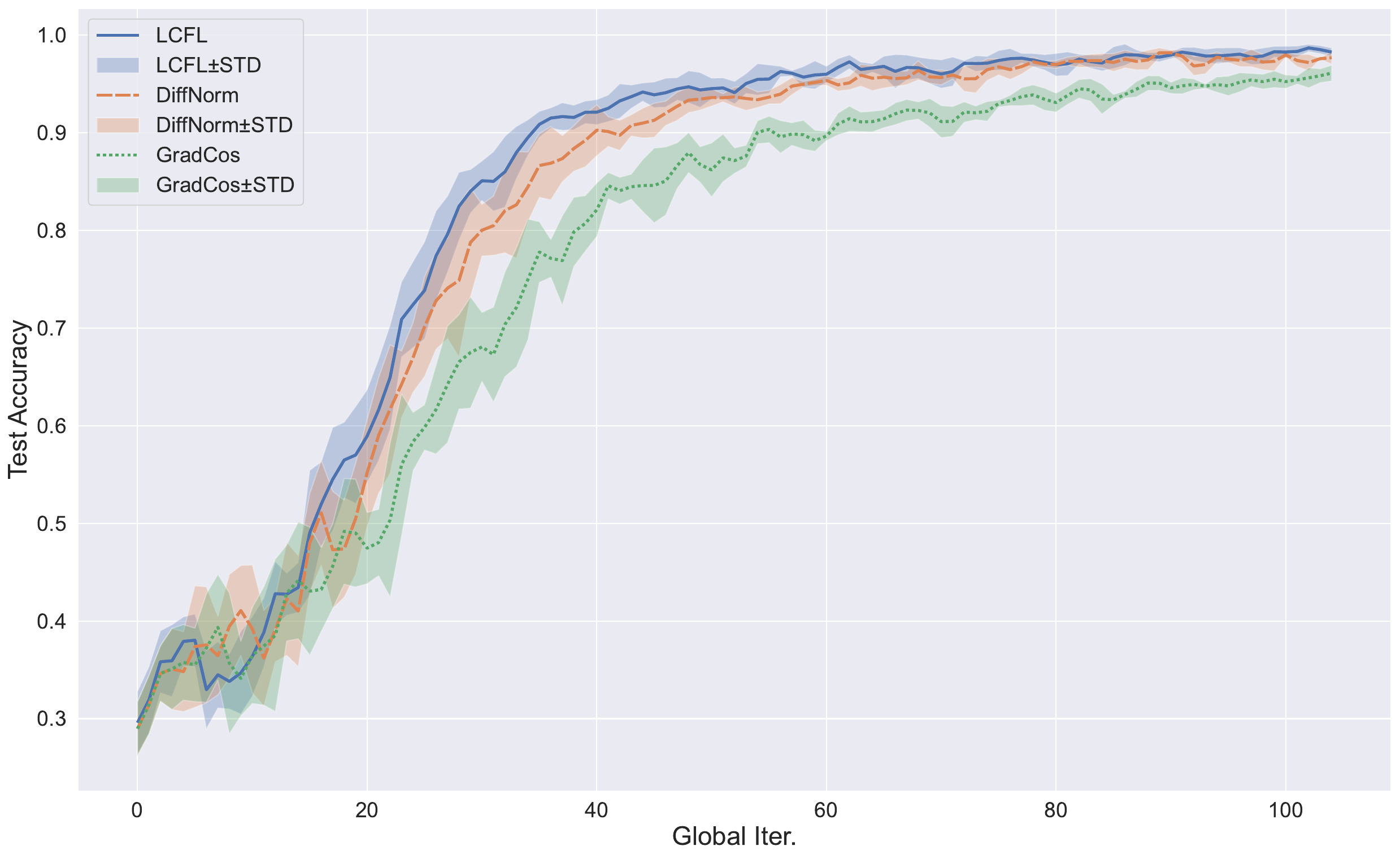}
   \caption{Performances of different metrics on FEMNIST}
   \label{fig: Performances of metrics on FEMNIST}
\end{figure}

\begin{figure}[htbp] 
   \centering
   \includegraphics[width=\linewidth]{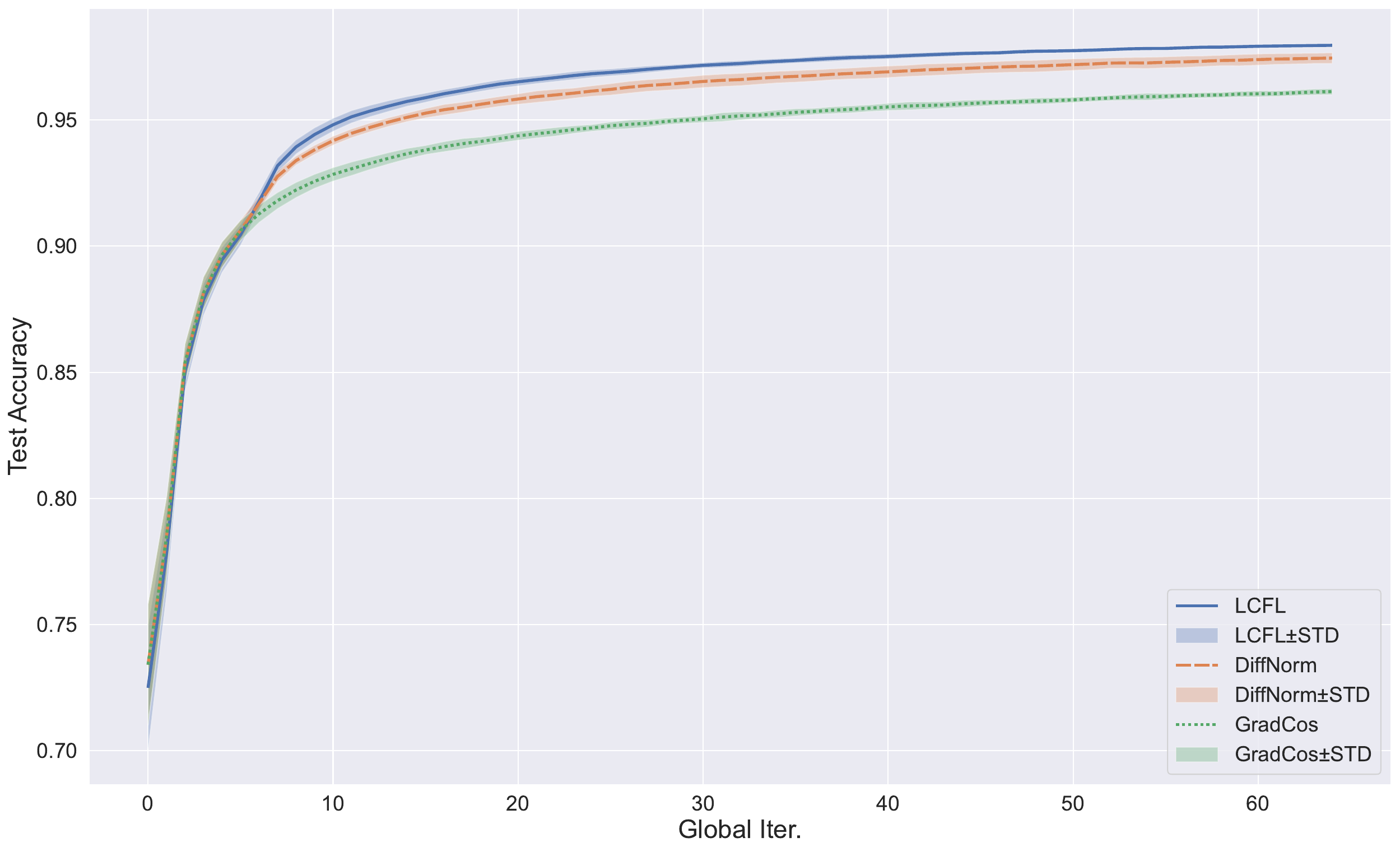}
   \caption{Performances of different metrics on Rotated MNIST}
   \label{fig: Performances of metrics on RotMNIST}
\end{figure}

The results are shown in Figure \ref{fig: Performances of metrics on FEMNIST} and Figure \ref{fig: Performances of metrics on RotMNIST}. In these figures, "LCFL" is our algorithm. "DiffNorm" and "Gradcos" refer to the metrics using model parameters and the cosine of gradients respectively. We conducted experiments for $m = 1200$ clients on Rotated MNIST. We repeated the experiment five times using five different random seeds and computed the standard deviation of the output accuracy with the "$\pm$STD" legend. We use average accuracy to measure the effect and "Global Iter." stands for the global iteration number.

In Figure \ref{fig: Performances of metrics on FEMNIST}, these three models ultimately achieved an average accuracy of over 90\%, but the number of iterations they took to achieve 90\% accuracy varied greatly. Our model achieved this result around the 37th iteration (considering standard deviation), while the other two measurement criteria were around the 40th iteration and the 60th iteration, respectively. On the results after 100 iterations, there is an approximate result of using the model loss functions and model parameters as metrics, which is about 98\%. However, using the cosine of gradients as the metric result is slightly worse, at around 95\%. The standard deviation of accuracy gradually decreases as the global iteration process progresses.

In Figure \ref{fig: Performances of metrics on RotMNIST}, the results are similar to the previous experiment shown in Figure \ref{fig: Performances of metrics on FEMNIST}, and using our proposed metric is superior to the other two metrics. In this experiment, the standard deviation of the average accuracy is very small. The clustering federated learning algorithm using three different metrics ultimately achieved an average accuracy of over 95\%, achieved in the 11th, 14th, and 27th iterations, respectively. After 60 iterations, the three metrics achieved results of approximately 96\%, 97\%, and 97.5\%, respectively.

The metric we proposed and utilized in the {\tt LCFL} algorithm is unambiguous, leading to the best performance compared to other metrics. This clear advantage is highlighted in the two figures. The unambiguous nature of our metric ensures that the clustering process in the {\tt LCFL} algorithm is robust and effective. By leveraging this metric, we can accurately capture the underlying patterns and similarities among the distributed data, resulting in improved clustering performance.

\subsection{Performances of algorithms}

In this subsection, we compare our proposed algorithm {\tt LCFL} with {\tt IFCA} and {\tt FedAvg}. Moreover, we conduct local training to verify the necessity of {\tt FL}. We conducted experiments on three datasets: FEMNIST, Rotated CIFAR10, and Rotated MNIST. For the experiments on Rotated MNIST, we conducted for $m=1200$ and $m=2400$ clients, respectively.

\begin{figure}[htbp] 
   \centering
   \includegraphics[width=\linewidth]{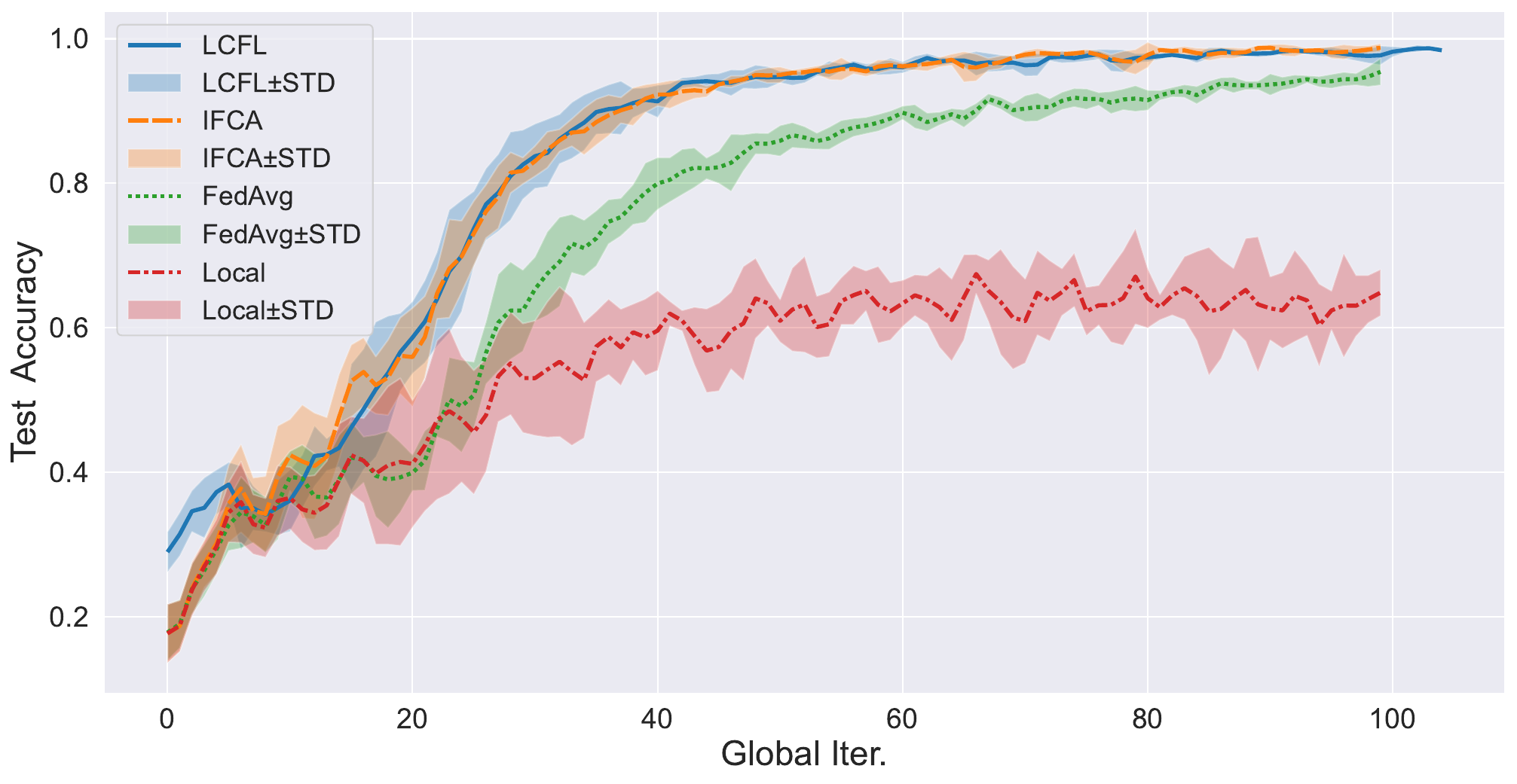}
   \caption{Test Accuracy comparison on FEMNIST}
   \label{fig:FEMNIST10-clusters10-val-acc}
\end{figure}

\begin{figure}[htbp] 
   \centering
   \includegraphics[width=\linewidth]{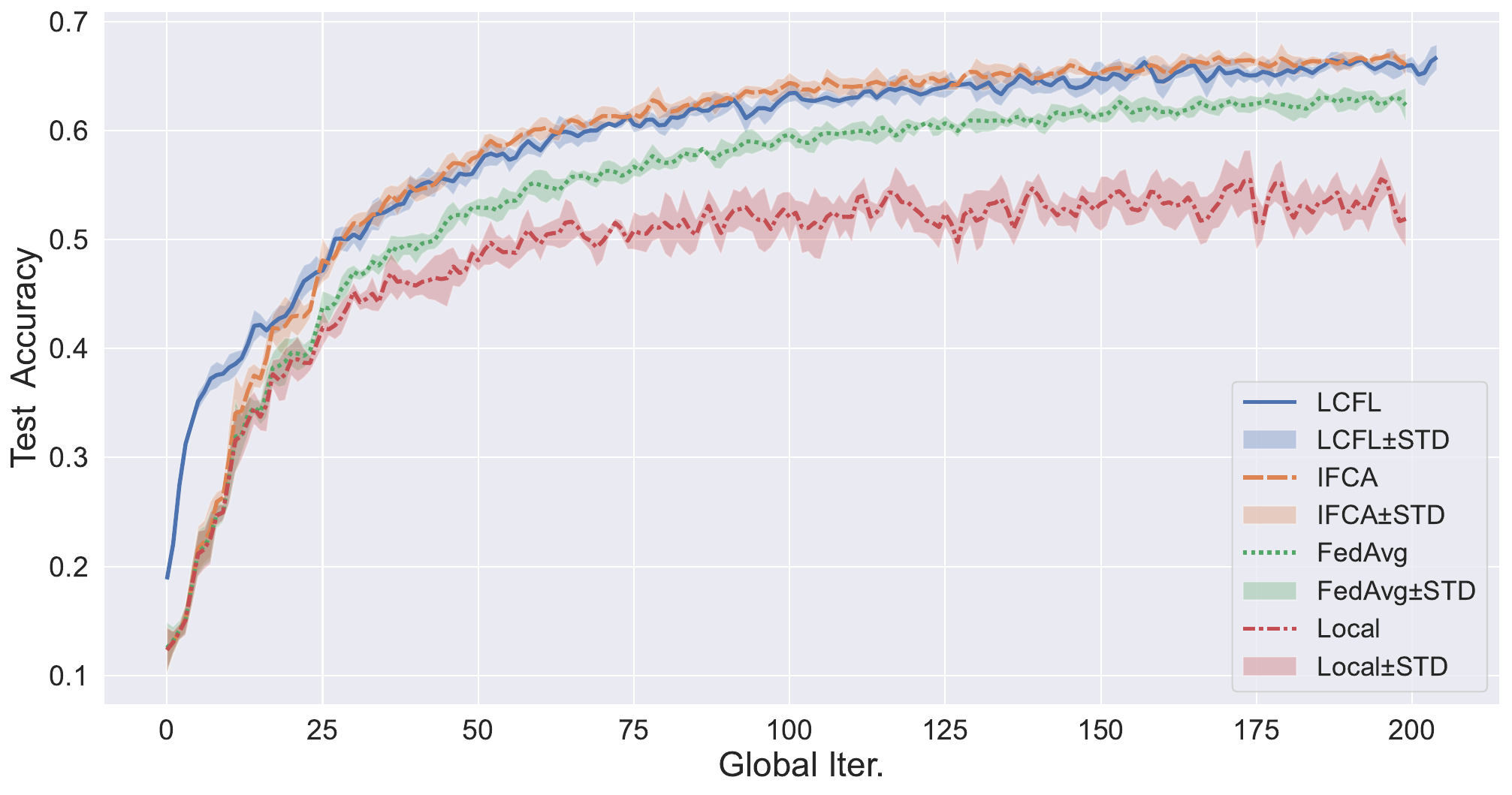}
   \caption{Test Accuracy comparison on Rotated CIFAR10}
   \label{fig:RotatedCIFAR10-dynamic-transform-val-acc}
\end{figure}

For {\tt LCFL}, we use the $k$-medoids to cluster users and set $k = 10, T = 10.$ For {\tt IFCA}, we also set $k = 10.$ For all algorithms, we run the experiments with 5 random seeds and report the average test accuracy and standard deviation in Figure \ref{fig:FEMNIST10-clusters10-val-acc}, Figure \ref{fig:RotatedCIFAR10-dynamic-transform-val-acc}, Table \ref{tab:Rotated MNIST 1200}, and Table \ref{tab:Rotated MNIST 2400}. The "Local" legend refers to the local training, calculating the average test set accuracy of all local models.

\begin{table}[htbp]
    \centering
    \caption{Test accuracies(\%) $\pm$ std(\%) on Rotated MNIST($m = 1200$)}
    \label{tab:Rotated MNIST 1200}
    \setlength{\tabcolsep}{1.2mm}{
    \begin{tabular}{|c|c|c|c|c|}
        \hlineB{3.5}
        Iteration & LCFL & IFCA & FedAvg & Local\\
        \hline \hline
        5  & $89.66 \pm 0.57$ & $91.13 \pm 0.51$ & $89.66 \pm 0.57$ & $74.82 \pm 1.86$\\
        10 & $94.45 \pm 0.31$ & $94.04 \pm 0.23$ & $92.52 \pm 0.31$ & $79.54 \pm 2.89$\\
        15 & $95.75 \pm 0.22$ & $95.14 \pm 0.19$ & $93.64 \pm 0.24$ & $84.33 \pm 3.22$\\
        30 & $97.11 \pm 0.12$ & $96.48 \pm 0.09$ & $95.03 \pm 0.17$ & $84.27 \pm 3.60$\\
        60 & $97.90 \pm 0.08$ & $97.42 \pm 0.09$ & $96.12 \pm 0.14$ & $84.33 \pm 3.75$\\
        \hlineB{3.5}
    \end{tabular}
    }
\end{table}

\begin{table}[htbp]
    \centering
    \caption{Test accuracies(\%) $\pm$ std(\%) on Rotated MNIST($m = 2400$)}
    \label{tab:Rotated MNIST 2400}
    \setlength{\tabcolsep}{1.2mm}{
    \begin{tabular}{|c|c|c|c|c|}
        \hlineB{3.5}
        Iteration & LCFL & IFCA & FedAvg & Local\\
        \hline \hline
        5  & $81.05 \pm 1.13$ & $85.48 \pm 1.39$ & $81.05 \pm 1.13$ & $78.66 \pm 1.34$\\
        10 & $91.05 \pm 1.11$ & $90.98 \pm 0.96$ & $87.51 \pm 0.50$ & $82.67 \pm 1.55$\\
        15 & $93.19 \pm 1.05$ & $92.77 \pm 0.82$ & $89.82 \pm 0.38$ & $84.12 \pm 1.67$\\
        30 & $95.30 \pm 0.84$ & $94.95 \pm 0.61$ & $92.53 \pm 0.27$ & $85.85 \pm 1.86$\\
        60 & $96.65 \pm 0.64$ & $96.41 \pm 0.50$ & $94.30 \pm 0.20$ & $87.01 \pm 1.96$\\
        \hlineB{3.5}
    \end{tabular}
    }
\end{table}

From the results of these experiments, our algorithm has demonstrated superior performance compared to the two baseline approaches, {\tt FedAvg} and local training, and is slightly better than {\tt IFCA}. By implementing the {\tt LCFL} algorithm, we can proactively uncover the underlying cluster identities of the clients. Once the correct cluster identity is discovered, each model is trained and tested using data from a similar distribution, resulting in improved accuracy. In contrast, the {\tt FedAvg} baseline attempts to fit all data from different distributions, which limits its ability to make personalized predictions, leading to inferior performance. The local model baseline algorithm is prone to overfitting, as proved by the gradually increasing standard deviation.

We can also see the advantages of clustering federated learning algorithms from our experimental results. As the test accuracy of {\tt FedAvg} after 60 global iterations is less than clustered {\tt FL} methods({\tt LCFL} and {\tt IFCA}).

Furthermore, the results provide compelling evidence for the necessity of conducting {\tt FL} tasks. They clearly illustrate that {\tt FL} models outperform local models by a significant margin. This emphasizes the importance of collaborative learning and leveraging the collective knowledge of distributed devices to enhance the performance and accuracy of machine learning models, demonstrating the superiority of {\tt FL} models over local models.

In summary, the clarity and effectiveness of our proposed metric, as well as the compelling evidence presented in the figures, support the superiority of {\tt FL} models over local models and emphasize the importance of conducting {\tt FL} tasks.



\section{Conclusion}
\label{sec:conclusion}

In this article, we presented {\tt LCFL}, a novel framework for clustered {\tt FL} that aims to improve existing {\tt FL} methods by enabling the participating clients to learn more specialized models through clustering. By leveraging our finding that the similarity between clients' loss functions is highly indicative of the similarity of their data distributions, {\tt LCFL} overcomes the limitations associated with using the similarity between weight updates.

To recap, {\tt LCFL} offers several advantages over previous clustered {\tt FL} approaches \cite{Sattler,Ghosh1,Ghosh2,Mansor,FPFC}. Firstly, it does not require the transfer of gradient data, making it particularly suitable for privacy-preserving situations. Secondly, {\tt LCFL} can effectively differentiate between scenarios where learning a single model from all clients' data is feasible and situations where it is not practical, and only segregates clients in the latter scenario. Moreover, the clustering step in {\tt LCFL} is performed as a warm-up, incorporating a classic {\tt FL} process.

Our experiment verified that using the value of loss functions as a metric has a better effect on clustering compared to the similarity between gradient cosine and that between weight update. In addition, our experiments on convolutional deep neural networks demonstrate that {\tt LCFL} can achieve significant advancements over the {\tt FL} baseline in terms of classification accuracy. This is particularly evident when all clients' data exhibit a clustering structure.

While this work has made significant strides in assessing the relationship between model performance and data distribution in FL, there are still areas for further investigation. Future research should delve into broader scenarios, conduct detailed analyses, and address potential limitations to provide a more comprehensive understanding of this field.

In conclusion, the {\tt LCFL} framework presents a promising approach for clustered {\tt FL}, with potential applications in various domains. By addressing privacy concerns, enhancing clustering accuracy, and improving classification performance, {\tt LCFL} contributes to the advancement of {\tt FL} methods and paves the way for more efficient and specialized learning in data-driven environments.



\section*{Acknowledgments}

This work is supported by NSFC under grants No. 12126603, and 12131004.

\bibliographystyle{IEEEtran}
\bibliography{bibfile}


\appendix

\begin{theorem}[Hoeffding's Inequality]
    Let $X_1, X_2, \ldots, X_n$ be independent and identically distributed (i.i.d.) random variables, where $X_i \in [a_i, b_i]$ for all $i$. Define the sample mean as $\bar{X}_n = \frac{1}{n}\sum_{i=1}^{n} X_i$. Then, for any $\epsilon > 0$, the following inequality holds:
    \begin{equation*}
        P\left(\left|\bar{X}_n - \mathbb{E}(\bar{X}_n)\right| \geq \epsilon\right) \leq 2e^{-2n\epsilon^2/\sum_{i=1}^{n} (b_i - a_i)^2}
    \end{equation*}
    where $\mathbb{E}(\bar{X}_n)$ denotes the expected value of $\bar{X}_n$.
\end{theorem}

\textbf{Proof of Theorem \ref{thm:bound}}
\begin{proof}
    The part to be bounded, denoted as $|d(i,j)-\hat{d}(i,j)|$, is decomposed into four terms by breaking down the absolute value:
    \begin{equation}\label{bound}
        \begin{aligned}
            & |d(i,j)-\hat{d}(i,j)| \\
            \leq & \ |L_{D_i}(w_j^\ast) -L_i(w_j^\ast)| + |L_i(w_i^\ast) - L_{D_i}(w_i^\ast)| \\
            & + |L_{D_j}(w_i^\ast) - L_j(w_i^\ast)| + |L_j(w_j^\ast) - L_{D_j}(w_j^\ast)|.
        \end{aligned}
    \end{equation}

    For the first part of the right hand of \eqref{bound}, the inequality is established using Hoeffding's inequality. By applying Hoeffding's inequality to the difference between the empirical and expected loss on samples drawn from $D_i,$ a lower bound is obtained. This bound is given by \eqref{proof1}.
    \begin{equation}\label{proof1}
        \mathop{\mathbb{P}}_{X_i\sim D_i^{m_i}} \left[ |L_{D_i}(w_j^\ast) - L_i(w_j^\ast)| \geq \epsilon \right] \leq 2\exp^{-2m_i\epsilon^2}.
    \end{equation}
    By setting the right-hand side of \ref{proof1} to be equal to $\delta$ and solving for $\epsilon,$ a new inequality is derived in the following:
    \begin{equation}\label{proof2}
        \mathop{\mathbb{P}}_{X_i\sim D_i^{m_i}} \left[ |L_{D_i}(w_j^\ast) - L_i(w_j^\ast)| \leq \sqrt{\frac{\log(2/\delta)}{2m_i}} \right] \geq 1 - \delta,
    \end{equation}
    Similarly, Hoeffding's inequality is applied to the difference between the empirical and expected loss on samples drawn from $D_j$, leading to the following two inequalities.
    \begin{equation*}
        \mathop{\mathbb{P}}_{X_j\sim D_j^{m_j}} \left[ |L_{D_j}(w_i^\ast) - L_j(w_i^\ast)| \geq \epsilon \right] \leq 2\exp^{-2m_j\epsilon^2},
    \end{equation*}
    and
    \begin{equation}\label{proof3}
        \mathop{\mathbb{P}}_{X_j\sim D_j^{m_j}} \left[ |L_{D_j}(w_i^\ast) - L_j(w_i^\ast)| \leq \sqrt{\frac{\log(2/\delta)}{2m_j}} \right] \geq 1 - \delta.
    \end{equation}
     According to Theorem 3.3 in \cite{mohri2018foundations}, the remaining two terms can be bounded using the generalization bounds in terms of Rademacher complexity. The probabilities of the differences between the true loss and expected loss being larger than a certain value are upper-bounded.
    \begin{equation*}
        \mathop{\mathbb{P}}_{X_i\sim D_i^{m_i}} \left[ |L_{D_i}(w_i^\ast) - L_i(w_i^\ast)| \geq \epsilon \right] \leq 2\exp^{-2m_i[\epsilon-\mathcal{R}_{m_i}(\mathcal{W})]^2},
    \end{equation*}
    \begin{equation*}
        \mathop{\mathbb{P}}_{X_j\sim D_j^{m_j}} \left[ |L_{D_j}(w_j^\ast) - L_j(w_j^\ast)| \geq \epsilon \right] \leq 2\exp^{-2m_j[\epsilon-\mathcal{R}_{m_j}(\mathcal{W})]^2},
    \end{equation*}
    We do the same transformation to get
    \begin{equation}\label{proof4}
        \begin{aligned}
            &\mathop{\mathbb{P}}_{X_i\sim D_i^{m_i}} \left[ |L_{D_i}(w_i^\ast) - L_i(w_i^\ast)| \leq \sqrt{\frac{\log(2/\delta)}{2m_i}} + \mathcal{R}_{m_i}(\mathcal{W}) \right] \\
            &\geq 1 - \delta
        \end{aligned}
    \end{equation}
    and
    \begin{equation}\label{proof5}
        \begin{aligned}
            &\mathop{\mathbb{P}}_{X_j\sim D_j^{m_j}} \left[ |L_{D_j}(w_j^\ast) - L_j(w_j^\ast)| \leq \sqrt{\frac{\log(2/\delta)}{2m_j}} + \mathcal{R}_{m_j}(\mathcal{W}) \right] \\
            &\geq 1 - \delta.
        \end{aligned}
    \end{equation}
    By combining the inequalities \eqref{bound}, \eqref{proof2},\eqref{proof3},\eqref{proof4} and \eqref{proof5}, the desired bound in \eqref{desire} is obtained. The probabilities of all the terms being within their respective bounds are multiplied, resulting in the final inequality.
    \begin{equation*}
        \begin{aligned}
            &\mathop{\mathbb{P}}_{\genfrac{}{}{0pt}{1}{X_i\sim D_i^{m_i}}{X_j\sim D_j^{m_j}}} \left[ |d(i,j)-\hat{d}(i,j)| \leq 2\sqrt{\frac{\log(2/\delta)}{2m_i}} +  2\sqrt{\frac{\log(2/\delta)}{2m_j}}\right. \\
            &\left. + \mathcal{R}_{m_i}(\mathcal{W}) + \mathcal{R}_{m_j}(\mathcal{W}) \right] \geq (1 - \delta)^4
        \end{aligned}
    \end{equation*}
    The final inequality guarantees that with probability at least $(1-\delta)^4$, the bound in \ref{desire} holds.
\end{proof}
Overall, this proof establishes a probabilistic bound on the difference between the target expected discrepancy $\hat{d}(i,j)$ and its empirical estimate $d(i,j)$, considering the individual terms and their respective probabilities.

\newpage

\section{Biography Section}

\begin{IEEEbiographynophoto}{Endong Gu}
received his B.S. degree in Statistics from Beihang University, Beijing, China, in 2021. Currently, he is pursuing his M.S. degree in Applied Mathematics at the School of Mathematical Sciences, Beihang University, Beijing, China. Endong's research interests primarily revolve around federated learning and the application of optimization problems in the field of machine learning.
\end{IEEEbiographynophoto}

\vspace{11pt}

\begin{IEEEbiographynophoto}{Yongxin Chen}
received his master's degree from the School of Mathematical Sciences, Nanjing Normal University. He is currently a Ph.D. candidate at the School of Mathematical Sciences, Beihang University, pursuing advanced studies in optimization theory and methods. His research interests include sparse optimization, distributed optimization, and nonconvex optimization.
\end{IEEEbiographynophoto}

\vspace{11pt}

\begin{IEEEbiography}[{\includegraphics[width=1in,height=1.25in,clip,keepaspectratio]{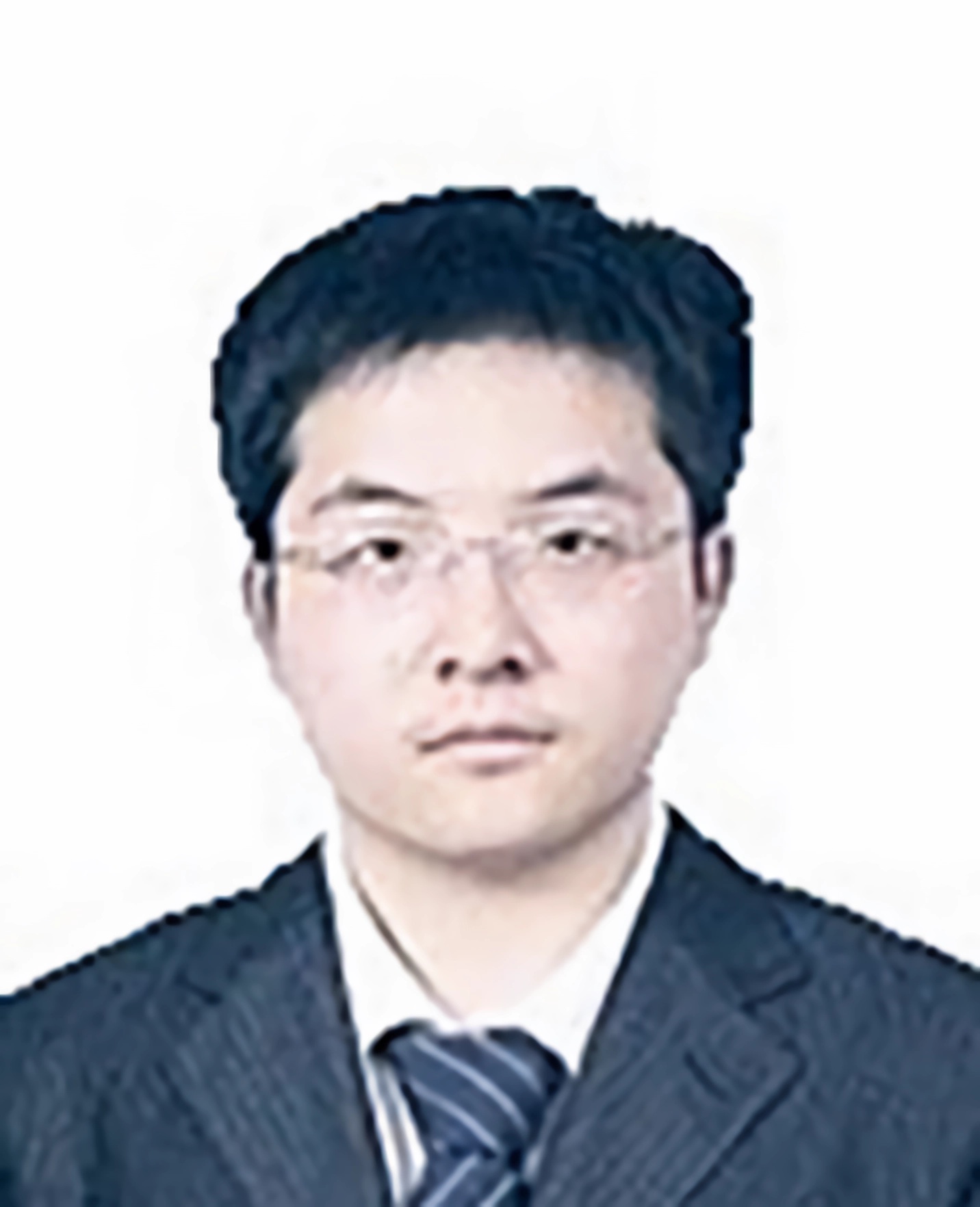}}]{Hao Wen}
received his Ph.D. degree in Mathematics in 2018 from Tsinghua University, Beijing, China. He worked as a post-doctoral fellow at Beihang University under the supervision of Professor Deren Han. He is currently a lecturer in the Department of Applied Mathematics in the College of Science at China Agricultural University. His primary research interests include theory, algorithm, and application of optimization; machine learning, especially federated learning, and its applications in medical and agricultural sciences.
\end{IEEEbiography}

\vspace{11pt}

\begin{IEEEbiographynophoto}{Xingju Cai}
is a Professor, and doctoral supervisor at the School of Mathematical Sciences, Nanjing Normal University. She is currently the deputy secretary-general of the Chinese Operations Research Society and the chairman of the Jiangsu Operations Research Society. She has served as a visiting editor of the Asia-Pacific Journal of Operational Research and Operations Research Transactions. She received her Ph.D. degree in Computational Mathematics from Nanjing University in 2013. Her research mainly focuses on machine learning, first-order algorithms for convex and non-convex optimization problems, and their computational complexity and applications. She has published over 30 academic papers in related fields and has made outstanding achievements in structural optimization and variational inequality decomposition algorithms. She was awarded the first prize for Science and Technology Progress of Jiangsu Province in 2021.
\end{IEEEbiographynophoto}

\vspace{11pt}

\begin{IEEEbiographynophoto}{Deren Han}
is a Professor, doctoral supervisor, the dean of the School of Mathematical Sciences, Beihang University. He is also the secretary general of the Mathematics Education Steering Committee of the Ministry of Education. He received his PhD degree in computational mathematics from Nanjing University in 2002. His research mainly focuses on numerical methods for large-scale optimization and variational inequality problems, as well as the application of optimization and variational inequality problems in transportation planning and magnetic resonance imaging. He has published many academic papers. He has won the Youth Operations Research Award of the Operations Research Society of China, the Second prize for Science and Technology Progress of Jiangsu Province, and other awards. He has presided over the National Science Fund for Distinguished Young Scholars and other projects. He serves as the standing director of the Operations Research Society of China and the president of the Operations Research Society of Jiangsu Province. Han also is an editorial board member of Numerical Computation and Computer Applications, the Journal of the Operations Research Society of China, Journal of Global Optimization.
\end{IEEEbiographynophoto}

\vfill

\end{document}